\documentclass{article}

\usepackage[preprint]{neurips_2024}

\usepackage[utf8]{inputenc} 
\usepackage[T1]{fontenc}    
\usepackage[hidelinks]{hyperref}       
\usepackage{url}            
\usepackage{booktabs}       
\usepackage{amsfonts}       
\usepackage{nicefrac}       
\usepackage{microtype}      
\usepackage[dvipsnames]{xcolor}         
\usepackage{xspace}
\usepackage{multirow}
\usepackage{listings}
\usepackage{color}
\usepackage{graphicx}
\usepackage{todonotes}

\definecolor{codegreen}{rgb}{0,0.6,0}
\definecolor{codegray}{rgb}{0.5,0.5,0.5}
\definecolor{codepurple}{rgb}{0.58,0,0.82}
\definecolor{backcolour}{rgb}{0.95,0.95,0.92}

\lstdefinestyle{mystyle}{
  backgroundcolor=\color{backcolour}, commentstyle=\color{codegreen},
  keywordstyle=\color{magenta},
  numberstyle=\tiny\color{codegray},
  stringstyle=\color{codepurple},
  basicstyle=\ttfamily\footnotesize,
  breakatwhitespace=false,         
  breaklines=true,                 
  captionpos=b,                    
  keepspaces=true,                 
  numbers=right,
  numbersep=5pt,                  
  showspaces=false,                
  showstringspaces=false,
  showtabs=false,                  
  tabsize=2,
}
\lstdefinestyle{added}{
  basicstyle=\ttfamily\color{green!50!black},
  backgroundcolor=\color{green!10},
  keywordstyle=\color{green!50!black},
}
\lstdefinelanguage{diff}{
  morecomment=[f][\color{green!50!black}]{\ },
  morecomment=[f][\color{red!50!black}]{\ },
}

\newcommand{\method}{MU-Bench\xspace}

\newcommand{\neggrad}{\textsc{NegGrad}\xspace}

\newcommand{\randlabel}{\textsc{RandLabel}\xspace}
\newcommand{\badt}{\textsc{Bad-T}\xspace}

\newcommand{\gnndel}{\textsc{GNNDelete}\xspace}
\newcommand{\scrub}{\textsc{SCRUB}\xspace}
\newcommand{\salun}{\textsc{SalUn}\xspace}

\newcommand{\dt}{$D_{\mathrm{Test}}$\xspace}
\newcommand{\df}{$D_f$\xspace}
\newcommand{\dr}{$D_r$\xspace}

\newcommand{\tl}{\textcolor{NavyBlue}{Loss}\xspace}
\newcommand{\rep}{\textcolor{YellowOrange}{Rep.}\xspace}
\newcommand{\out}{\textcolor{BrickRed}{Logit}\xspace}

\newcommand{\cd}{\textcolor{NavyBlue}{Data}\xspace}
\newcommand{\cg}{\textcolor{YellowOrange}{Grad}\xspace}
\newcommand{\cm}{\textcolor{BrickRed}{Model}\xspace}

\newcommand{\pd}{\textcolor{BrickRed}{Dense}\xspace}
\newcommand{\ps}{\textcolor{NavyBlue}{Sparse}\xspace}

\newcommand{\pin}{\textcolor{ForestGreen}{Internal}\xspace}
\newcommand{\pex}{\textcolor{YellowOrange}{External}\xspace}

\title{\method: A Multitask Multimodal Benchmark\\for Machine Unlearning}

\author{%
  Jiali Cheng \quad Hadi Amiri \\
  University of Massachusetts Lowell\\
  \texttt{\{jiali\_cheng, hadi\_amiri\}@uml.edu} \\
}

\begin{document}

\maketitle

\begin{abstract}
Recent advancements in Machine Unlearning (MU) have introduced solutions to selectively remove certain training samples, such as those with outdated or sensitive information, from trained models. Despite these advancements, evaluation of MU methods have been inconsistent, employing different trained models and architectures, and sample removal strategies, which hampers accurate comparison. In addition, prior MU approaches have mainly focused on {\em singular} tasks or modalities, which is not comprehensive. To address these limitations, we develop \method, the first comprehensive benchmark for MU that 
\emph{(i) unifies the sets of deleted samples and trained models}, and
\emph{(ii) provides broad coverage of tasks and data modalities}, 
including previously unexplored domains such as speech and video classification. 
Our evaluation show that \randlabel~\citep{amnesiac_2021} and \salun~\citep{fan2024salun} are the most effective general MU approaches on \method, and \badt~\citep{Chundawat2022CanBT} and \scrub~\citep{unbound} are capable of achieving random performance on the deletion set. 
We analyze several under-investigated aspects of unlearning, including scalability, the impacts of parameter-efficient fine-tuning and curriculum learning, and susceptibility to dataset biases. 
\method provides an easy-to-use package that includes dataset splits, models, and implementations, together with a leader board to enable unified and scalable MU research.\footnote{Project page: \url{https://clu-uml.github.io/MU-Bench-Project-Page}.}.

\end{abstract}
\section{Introduction}
Machine Unlearning (MU) aims at selectively removing a small portion of training data--and the influence of the samples--from a trained model. MU is essential for protecting sensitive information and discarding outdated samples. Recent works have studied machine unlearning in various contexts, including classification tasks on image~\citep{guo-2020-certified-removal,tang-etal-2023-boundary} and graph~\citep{Chien2023EfficientMU,cheng2023gnndelete} data, 
multimodal tasks~\citep{cheng2023multimodal}, 
generation tasks~\citep{chen-yang-2023-unlearn,Gandikota_2023_ICCV,fan2024salun}, and
federated learning~\citep{wang2021federated}.

Despite these advancements, existing approaches to machine unlearning face several challenges: 
(1): MU systems are evaluated under inconsistent settings, using different trained models (from which data is deleted) and metrics, which can lead to unfair comparisons and hinder the development of robust unlearning approaches~\citep{fan2024salun}; 
(2): evaluation tend to focus on specific tasks, modalities, and architectures, which limits our understanding on the effectiveness of these models across different settings~\citep{wang-etal-2023-kga,Chundawat2022CanBT}.

To address these limitations, we introduce \method, a comprehensive machine unlearning benchmark consisting of multiple tasks, data modalities, base models, standardized evaluation metrics, all compiled into an easy-to-use package with a leader board to enable robust and scalable MU research.   
To the best of our knowledge, this benchmark represents the first effort to benchmark existing MU approaches across a wide range of settings.

Our contributions are:
\begin{itemize}
\itemsep0pt
    \item constructing the first comprehensive MU benchmark with a wide coverage of tasks, domains, and modalities, including previously unexplored areas, such as speech and video processing, and biomedical applications, for systematic evaluation of unlearning algorithms;
    
    \item unifying (and perhaps democratizing) MU with uniformed deleted samples and a wide range of trained models and architectures 
    to enable fair comparisons between MU methods; 
    
    \item identifying design choices that explain performance variations across tasks and modalities;
    
    
    \item investigating several overlooked aspects of unlearning, such as deletion capacity, parameter-efficient fine-tuning (PEFT), and the impact of curriculum learning and dataset bias to inform future research directions.

    
    
\end{itemize}

Extensive experiments show that \randlabel~\citep{amnesiac_2021}, \badt~\citep{Chundawat2022CanBT}, and \salun~\citep{fan2024salun} are generally robust MU methods. When operating under a fixed training budget of compute (FLOS), \randlabel and \salun outperform \badt. We find that existing MU methods benefit from PEFT but much less than other learning tasks, where below is 50\% of the entire parameters, below which the model cannot be trained. 
Moreover, Curriculum Learning techniques can help models forget less and does not facilitate MU in most cases.
In addition, performance variations across different tasks and modalities suggest that specific design choices within MU approaches significantly influence their effectiveness. In particular, certain tasks such as audio and video classification, are challenging for existing MU methods.

By design, \method is structured to incorporate new datasets and tasks, and we will continue to expand its resources in future. 



\begin{figure}
  \centering
  \vspace{-3em}
  \includegraphics[width=\textwidth]{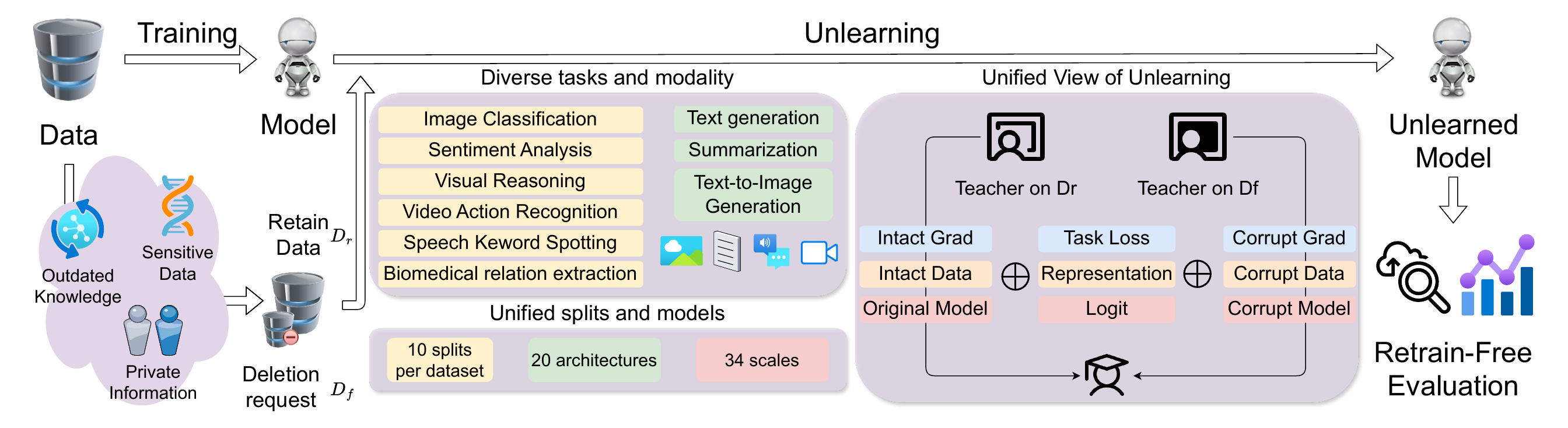}
  \caption{The \method benchmark for machine unlearning (MU) spans a comprehensive range of tasks and modalities, including previously unexplored data types such as audio, video, and biomedical data. The open-source package of \method provides standardized (unified) data splits, implements a suite of commonly-used MU methods and their design choices, enables fast experimentation and fair comparisons across MU methods, and is structured to easily incorporate new datasets and tasks in future.\looseness-1}
\end{figure}

\section{\method}
We outline the design of \method, covering tasks, datasets, models, and evaluation metrics.

\subsection{Problem Formulation}

\paragraph{Machine unlearning} Let $D_{\mathrm{Train}}$ denotes the training dataset, $D_f \subseteq D_{\mathrm{Train}}$ the subset to be unlearned, and $D_r = D_{\mathrm{Train}} \backslash D_f$ the remaining dataset post-unlearning. Given a model $f$ trained on $D_{\mathrm{Train}}$, machine unlearning seeks to remove the influence of $D_f$ from $f$ without affecting the knowledge it gained from $D_r$, without retraining from scratch. We term $f$ as the original model and $f'$ as the model post-unlearning. A successful unlearned model $f'$ should be minimally impacted by $D_f$, 
while maintaining the performance of $f$ on the original downstream test set \dt.

\paragraph{Evaluation Metrics}
Evaluating the efficacy of unlearning is crucial for identifying models that are more secure and retain no/less memory of deleted data. While previous studies have employed different metrics, we propose a set of metrics that do not require model retraining: 
performance on test set \dt ($\uparrow$), 
performance on deletion set \df ($\downarrow$), 
performance on remaining set \dr ($\uparrow$), 
unlearning time ($\downarrow$), and 
success rate of membership inference attack ($\downarrow$).

\paragraph{Toward a retrain-free evaluation} Early works in machine unlearning research often considered the model retrained from scratch on \dr as the {\em gold} standard for $f'$, which is now recognized as an inappropriate design choice due to several issues:
\textbf{First}, evaluating $f'$ based solely on its closeness or similarity to the retrained model can lead to false negatives. This is because the parameters of $f'$ may fall onto different distributions than the retrained model, but still achieve competitive unlearning performance. On the other hand, the parameters of two models can match even with completely different training datasets~\citep{lamproudis-etal-2022-evaluating}. 
\textbf{Second}, retrained models cannot guarantee the privacy of deleted data in practice, often maintaining undesired high performance on \df, as demonstrated by previous work~\citep{cheng2023gnndelete}. 
\textbf{Third}, obtaining a precise \dr can be impractical in cases where the goal of unlearning is to remove toxic content~\citep{NEURIPS2023_299a08ee,ilharco2023editing} or abstract concepts~\citep{Gandikota_2023_ICCV}. Such abstract concepts may not correspond to identifiable data samples. 
\textbf{Finally}, retraining a model from scratch on $D_r$ can be impractical or even impossible due to confidentiality constraints, 
proprietary data concerns, or because the data may no longer be available. In addition, retraining is often expensive, especially for large datasets or complex tasks such as multimodal learning or large language models (LLMs). 
Based on the above shortcomings, we advocate for a retrain-free evaluation of unlearning systems, a method that is increasingly recognized in recent works~\citep{Chundawat2022CanBT}.



\begin{table}[t]
\caption{Example datasets currently available in \method, covering a wide set of tasks and data modalities from different domains. 
$|D|$ denotes the size of training data. In \method, we set the deletion ratio to a maximum of 10\% of $|D|$. Rows labeled with \textbf{*} indicate new tasks and data modalities introduced in \method for machine unlearning.}
\label{tab:data}
\centering
\small
\begin{tabular}{llp{5.5em}lp{1.5em}}
\textbf{Dataset} & \textbf{Task} & \textbf{Domain} & \textbf{Modality} & $\mathbf{|D|}$ \\
\toprule
\multicolumn{5}{c}{\textbf{Discriminative Tasks}}\\
\midrule
CIFAR-100~\tiny{\citep{cifar100}} & Image classification & General & Image & 50K \\
IMDB~\tiny{\citep{maas-etal-2011-learning}} & Sentiment classification & Movie review & Text & 25K \\
\textbf{*} DDI-2013~\tiny{\citep{segura-bedmar-etal-2013-semeval}} & Relation extraction & Biomedical & Text & 25K \\
NLVR\textsuperscript{2}~\tiny{\citep{suhr-etal-2019-corpus}} & Visual reasoning & General & Image-Image-Text & 62K \\
\textbf{*} Speech Commands~\tiny{\citep{warden2018speech}} & Keyword spotting & Commands & Speech & 85K \\
\textbf{*} UCF101~\tiny{\citep{soomro2012ucf101}} & Action classification & General & Video & 9.3K \\
\midrule
\multicolumn{5}{c}{\textbf{Generative Tasks}}\\
\midrule
SAMSum~\tiny{\citep{gliwa-etal-2019-samsum}} & Text summarization & Chat dialogue & Text & 14K \\
\textbf{*} BioFact~\tiny{\citep{min-etal-2023-factscore}} & Text generation & Biography & Text & 183 \\
Tiny ImageNet~\tiny{\citep{tinyimagenet}} & Text-to-Image generation & General & Image-Text & 20K \\

\bottomrule
\end{tabular}
\end{table}

\subsection{Datasets and Tasks}
We adopt nine publicly available datasets covering a diverse set of discriminative and generative tasks and data modalities.  As Table~\ref{tab:data} shows, the discriminative tasks include
CIFAR-100~\citep{cifar100} for image classification, 
IMDB~\citep{maas-etal-2011-learning} for sentiment classification, 
DDI~\citep{segura-bedmar-etal-2013-semeval} for relation extraction in the biomedical domain, 
NLVR2~\citep{suhr-etal-2019-corpus} for visual reasoning, 
Speech Commands~\citep{warden2018speech} for keyword spotting, and
UCF101~\citep{soomro2012ucf101} for action classification.
The generative tasks include 
SAMSum~\citep{gliwa-etal-2019-samsum} for text summarization, 
Biography (adapted from \citet{min-etal-2023-factscore}, see below) for text generation,
Tiny ImageNet~\citep{tinyimagenet} for text-to-image generation.

We build a new dataset for evaluating machine unlearning in large language models (LLMs), focusing on the removal of personal information, as a common unlearning request. This is a crucial tasks because for example, on social media, user can choose to delete their accounts or privatize them, resulting in a critical and perhaps legal impetus for machine unlearning. The dataset contains factual descriptions of 183 celebrities, obtained from~\citep{min-etal-2023-factscore}, to enable machine unlearning of personal data from LLMs. 

These datasets were chosen for their relevance  to practical machine unlearning tasks, their variety, including both well-established and under-explored datasets, and their capacity to highlight differences between unlearning methods across diverse tasks and modalities (as they have non-saturated performance). 
This datasets allow for large scale and fair evaluation of unlearning methods, and addresses gaps in current research in several unexplored areas in machine unlearning.

\begin{lstlisting}[basicstyle=\tiny,label=lst:example,language=Python, caption=Example usage of \method: Unlearning 5\% data from the VILT model trained on NLVR2 using MultiDelete.]
# Standard HuggingFace code
from transformers import TrainingArguments
args = TrainingArguments(output_dir="tmp")

# Additional code for unlearning
from mubench import UnlearningArguments, unlearn_trainer
unlearn_config = UnlearningArguments(
    unlearn_method="multi_delete",  # MU method, MultiDelete ECCV'24
    backbone="vilt",                # Network architecture
    data_name="nlvr2",              # Dataset
    del_ratio=5                     # Standardized splits
)
trainer = unlearn_trainer(unlearn_config.unlearn_method)(
    args=args, 
    unlearn_config=unlearn_config
)
trainer.unlearn()                   # Start Unlearning and Evaluation!

\end{lstlisting}

\subsection{Unified Unlearning}
To address inconsistencies in the evaluation of MU approaches, we unify critical aspects such as \emph{the choice and size of deleted samples} (\df), and \emph{the baseline model ($f$)} from which data is removed. This unification allows for meaningful comparison and democratizes access through open-source tools.\looseness-1

\paragraph{Deleted Samples} 
For each dataset, we randomly sample 1-10\% of the training data as \df, with increments of 1\% to covers both typical and extreme evaluation settings. This approach reflects typical and realistic settings where a small portion of data is deleted~\citep{Golatkar2020EternalSO,Chundawat2022CanBT,cheng2023gnndelete}, and challenges the limits of unlearning methods without fundamentally altering the data distribution, as would be the case with more extensive data removal.


\paragraph{Original Model}
For each dataset, we train a set of commonly-used models on different architectures and scales, from which \df is deleted, to allow for robust and relevant comparisons. We train a total of 20 architectures and 34 scales, such as
ResNet~\citep{He_2016_CVPR} (18, 34, 50 layers), 
ViT~\citep{dosovitskiy2021an} (Small, Base, Large), 
Swin-Transformer~\citep{Liu_2021_ICCV} (Tiny, Small, Base), 
MobilNet V2~\citep{Sandler_2018_CVPR} for image classification; and HuBERT~\citep{hubert} (Base, Large, X-Large), 
Whisper~\citep{pmlr-v202-radford23a} (Tiny, Small, Base), 
Wav2Vec2.0~\citep{NEURIPS2020_92d1e1eb} (Base, Large) for the audio classification. 
Additional details are provided in Appendix~\ref{sec:ori}.

\paragraph{Example Usage}
We include the datasets, standardized data splits, evaluation scripts, and unlearning methods within an easy-to-use Python package and integrate them with commonly-used packages such as PyTorch~\citep{pytorch}, Huggingface Transformers~\citep{wolf-etal-2020-transformers}, and Diffusers~\citep{von-platen-etal-2022-diffusers}, containing pre-trained diffusion models for image and speech data. Users can initiate an unlearning experiment with minimal adjustment to existing script. All original model checkpoints are released for standardized unlearning and fair comparisons. We also host and maintain a leaderboard to rank methods overall and on individual tasks and architectures.
For example, to remove 5\% of training data from a BERT-base model trained on IMDB using \badt~\citep{Chundawat2022CanBT}, only a minimal script modification is required shown in code example~\ref{lst:example}. 
This setup simplifies the unlearning process and enables rapid comparison against methods and architectures.\looseness-1



\paragraph{Taxonomy of Unlearning Techniques: A Teacher-Student Framework}
To provide a deeper understanding of the design choices of existing MU approaches and their performance differences, we introduce a taxonomy based on a unified teacher-student framework. In this framework, the desired unlearned model $f'$ seeks to selectively discard specific knowledge from the original model $f$ under the guidance of a ``teacher.' As shown in Table~\ref{tab:unified_view}, the design choices of the teacher vary across different methods mainly from three aspects:
\begin{itemize}
\itemsep-1pt
    \item \textbf{Knowledge Measurement (KM)}: the key question of how knowledge is quantified, which is determined by task loss (\tl), representation (\rep), or output logits (\out) in existing MU models;
    
    \item \textbf{Knowledge Corruption on $D_f$ (Corrupt)}: the key question of how the knowledge associated with $D_f$ is degraded, which is currently determined using techniques such as reversing gradients (\neggrad), using random data (\randlabel), or employing an incompetent teacher (\badt); and
    
    \item \textbf{Knowledge Retention on $D_r$ (Retain)}: the key question of how to preserve knowledge from $D_r$, which is typically achieved by treating the original model $f$ as the teacher. 
\end{itemize}
These elements combine differently across methods, influencing both the teacher's role on $D_f$ and $D_r$, as detailed in Table~\ref{tab:unified_view}; specifically, (i) and (ii) lead to teacher on $D_f$, and (ii) and (iii) lead to teacher on $D_r$. 
In Addition, the trainable parameters can be dense or sparse and internal or external.
We utilize this taxonomy to categorize common and distinctive design elements in existing methods. This categorization helps in understanding how different unlearning approaches function and enables their transfer and adaptation to new contexts, such as generative tasks.

\paragraph{Extension to generative tasks}
Even though many unlearning methods are designed for and evaluated on classification tasks, they can be applied to generative tasks with minimal modifications. For example, in case of \randlabel, data pairs $(x, y)\in D_f$ can be altered to $(x, y')$ where $y' \in D_r, y' \neq y$. For \badt, the method can be adjusted to match the predictions of each token when measuring the teacher-student divergence. 

\begin{table}[t]
\caption{Taxonomy of unlearning techniques. Despite different formulations and loss functions, existing approaches can be viewed in a unified teacher-student framework, with three design choices: (i) knowledge measurement (KM), (ii) knowledge corruption on $D_f$ (Corrupt), and (iii) knowledge retention on $D_r$ (Retain). The combination of (i)  and (ii) leads to teacher on $D_f$, while combination of (i) and (iii) leads to teacher on $D_r$.
For teachers on $D_f$ and $D_r$, \tl represents the expected task loss $\mathbb{E}_{(x, y) \in D}\sum L(f(x), y)$ on $D_f$ and $D_r$. 
\rep denotes the KL Divergence of output distribution $\mathbb{E}_{(x, y) \in D}\sum \mathrm{KL}(f'(x), f(x))$ on $D_f$ and $D_r$. 
%
%
Trainable parameters are denoted as \pd or \ps, and \pin or \pex.} 
\label{tab:unified_view}
\small
\centering
\begin{tabular}{p{4.5cm}|p{1.5cm}p{1cm}|p{1.5cm}p{1cm}|c}
    \multirow{2}{*}{\textbf{Method}} & \multicolumn{2}{c|}{\textbf{Teacher on }$D_f$} & \multicolumn{2}{c|}{\textbf{Teacher on }$D_r$} & \multirow{2}{*}{\textbf{Parameters}} \\ 
                            & \textbf{KM} & \textbf{Corrupt} & \textbf{KM} & \textbf{Retain }& \\
    \toprule
    Exact unlearning                        & -- & -- & \tl & $f$                                       & \pd, \pin \\
    \midrule
    \neggrad~\citep{Golatkar2020EternalSO}  & \tl        & \cg  & --         & --   & \pd, \pin \\
    \randlabel~\citep{amnesiac_2021}        & \tl        & \cd  & \tl        & $f$  & \pd, \pin \\
    \badt~\citep{Chundawat2022CanBT}        & \out       & \cm  & \out       & $f$  & \pd, \pin \\
    \scrub~\citep{unbound}                  & \tl        & \cg  & \tl + \rep & $f$  & \pd, \pin \\
    \salun~\citep{fan2024salun}             & \tl        & \cd  & \tl        & $f$  & \ps, \pin \\
    $l_1$-sparse FT~\citep{jia2023model}    & --         & --   & \tl        & $f$  & \ps, \pin \\
    MultiDelete~\citep{cheng2023multimodal} & \rep       & \cd  & \rep       & $f$  & \pd, \pin \\
    EUL~\citep{chen-yang-2023-unlearn}      & \tl + \rep & \cg  & \tl + \rep & $f$  & \pd, \pex \\
    UL~\citep{jang-etal-2023-knowledge}     & \tl        & \cg  & --         & --   & \pd, \pin \\
    \gnndel~\citep{cheng2023gnndelete}      & \rep       & \cd  & \tl + \rep & $f$  & \pd, \pex \\
    SGA-TAU~\citep{barbulescu2024textual}   & \tl        & \cg  & --         & --   & \ps, \pin \\
    \bottomrule
\end{tabular}
\end{table}

\section{Experiments}

\paragraph{Setup}
For each dataset, we first train the task-specific original model $f$ long enough with hyperparameter optimization and select the best performing model. This is usually the practice for models deployed for real world applications. For LLM and Text-to-Image generation tasks, we evaluate unlearning from the pretrained models, since they are not fine-tuned for a specific task. In addition, we limit the unlearning time so that it does not exceed the retraining time, otherwise unlearning would not be practical. We repeat all experiments five times with different random seeds to account for stochastic effects. We focus on the following MU models selected based their widespread usage and unique characteristics: \neggrad~\citep{Golatkar2020EternalSO}, \randlabel~\citep{amnesiac_2021}, \badt~\citep{Chundawat2022CanBT}, \scrub~\citep{unbound}, and \salun~\citep{fan2024salun}. Details on the architectures used can be found in~\ref{sec:ori} and the performance of the other MU models will be available on the leaderboard.

\subsection{Main Results on Discriminative Tasks}

\begin{figure}[t]
  \centering
  \vspace{-20pt}
  \includegraphics[width=\textwidth]{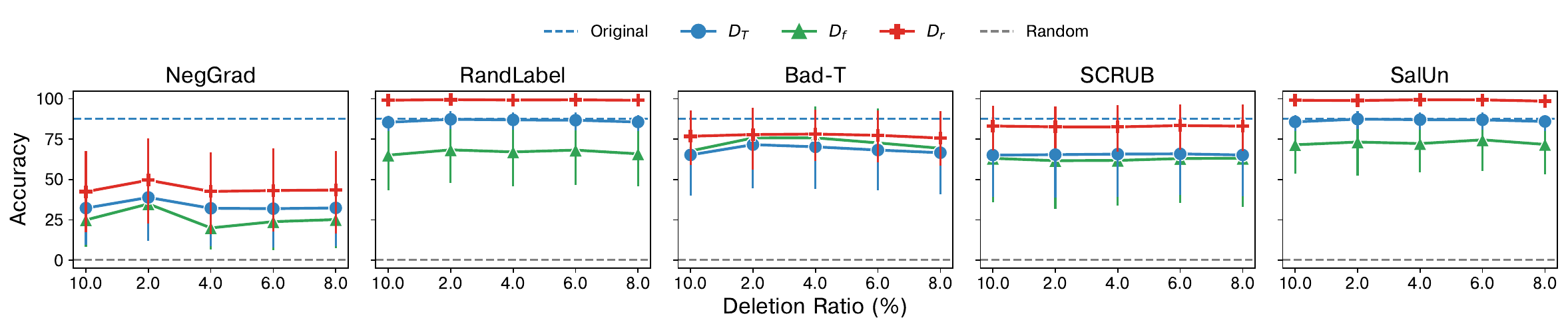}
  \vspace{-10pt}
  \caption{Overall average accuracy across all discriminative tasks.}
  \label{fig:ave_cls_perf}
\end{figure}

As Figure~\ref{fig:ave_cls_perf} illustrates, \neggrad typically results in low performance on \df, but severely compromises the knowledge on \dt and \dr, indicating it is not an effective MU approach. In general, tasks like audio classification, video classification, text summarization and generation consistently challenge existing MU algorithms, potentially due to strong correlations within the data, see Figure~\ref{fig:superb_ks_perf}-\ref{fig:ucf101_perf}). We report the average performance across all tasks as all metrics range from 0 to 100\%.

For image classification on CIFAR-100, \badt achieves close-to-random performance on \df while preserving 40\% accuracy on \dt and \dr. Both \randlabel and \salun effectively maintain models' capability on downstream test sets but fails to forget the deletion set. The original \salun paper reported slightly different results, which we hypothesize may be due to the class-balanced sampling strategy and nuanced class hierarchy of CIFAR-100. Interestingly, \scrub achieves very similar performances on \dt, \df, and \dr, see Figure~\ref{fig:cifar100_perf}. 

For sentiment classification on IMDB, \randlabel and \salun show promising results in forget \df with close-to-random performances, with minimal impact on \dt. \badt and \scrub also preserve strong performance on \dt but fail to unlearn \df. Since IMDB contains strong dataset biases and shortcut features, corrupting the data labels implemented by \randlabel and \salun seems to be a more effective approach than corrupting gradient, see Figure~\ref{fig:imdb_perf}.

For biomedical relation extraction on DDI, \randlabel, \scrub, \salun all succeed in 
forgetting the deletion set \df, with \scrub slightly impairing test performance more than others. Conversely, \badt completely failed to unlearn \df, see Figure~\ref{fig:ddi_perf}. 

For visual reasoning on NLVR2, \randlabel and \salun again are successful in unlearning \df, unlike \badt and \scrub, which failed to unlearn \df. However, one potential issue with \salun is in the excessively low performance, almost close to zero performance, on \df, which may be too low and prone to information leakage. This will be further discussed in \S\ref{sec:lowdf}, see Figure~\ref{fig:nlvr2_perf}. 

The speech keyword spotting on Speech Commands show that none of the existing methods can forget \df without severely impacting knowledge retention. Either with minimal knowledge removed (\neggrad, \randlabel, \salun), or resulting in too much performance degradation on \dt (\badt, \scrub). This can potentially be due to the correlations between audio waves, for which prior approaches do not have mechanisms to handle, see Figure~\ref{fig:superb_ks_perf}.

For video action recognition on UCF101, all methods maintain original performance on \dt and \dr, but all fail to forget \df, with 90+\% accuracy. This can be attributed to the fact that current video classification methods rely on inter-frame correlation, while existing MU methods lacks mechanisms to remove such information, leading to failed unlearning, see Figure~\ref{fig:ucf101_perf}.


\begin{figure}[t]
  \centering
  \vspace{-10pt}
  \includegraphics[width=\textwidth]{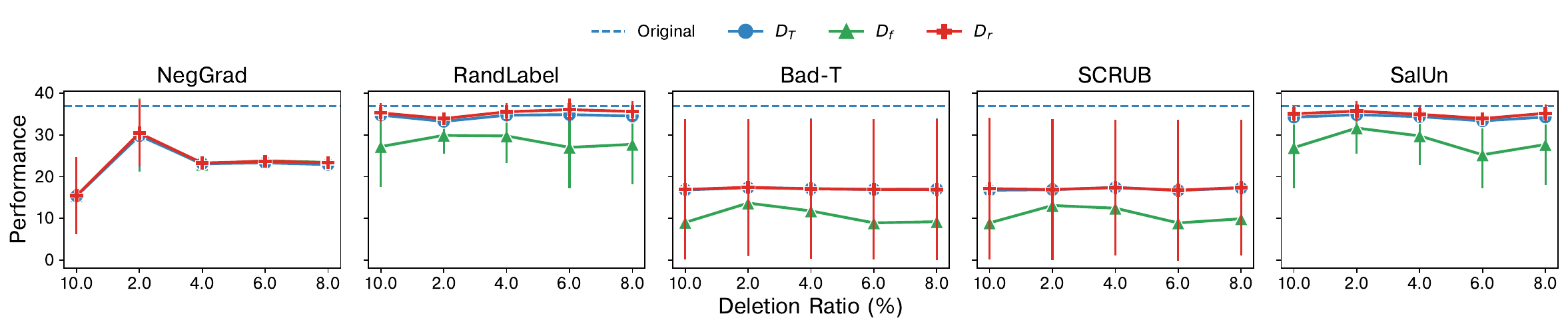}
  \vspace{-20pt}
  \caption{Overall average performance across all generative tasks.}
  \label{fig:ave_gen_perf}
\end{figure}

\subsection{Main Results on Generative Tasks}

In general, generation tasks present greater challenges for unlearning and evaluation. As Figure~\ref{fig:ave_gen_perf} shows, for text summarization on SAMSum and text generation on BioFact, existing general MU approaches all fail to achieve unlearning. \randlabel and \salun has limited influence over all data including \df and \dr, while \badt and \scrub remove knowledge of all data. 
In addition, we find that \neggrad show very different performance pattern on generative tasks compared to discriminative tasks, with non-random performance when a small portion of examples are deleted, see Figure~\ref{fig:samsum_perf}. 


For text-to-image generation, we find all methods can effectively reduce the clip score between image-prompt pairs on \df with limited impact on \dt and \dr, see Figure~\ref{fig:tiny_imagenet_perf}). To ensure the generated images are not from the orginal classes, we use a trained image classifier to classify the samples in \df. \salun outperforms all other approaches by 5.1 in accuracy on average, see Table~\ref{tab:image_gen_acc}.




Additional results on training time and membership inference attack are shown in Appendix~\ref{sec:time}.

\section{Discussion and Analysis}

\paragraph{What is the deletion capacity of each method?} We define {\em deletion capacity} as the amount of data a model can forget without degrading performance on $D_T$. \randlabel and \salun have relatively larger deletion capacity than \scrub, while \badt has the smallest capacity. These results suggest that task loss is a potentially better way of knowledge measurement than matching logits in \badt. Another reason is the computation cost of \badt restricts its capability of forgetting more samples. 
Furthermore, we find that the deletion capacity of the same MU method varies across different tasks, modalities, and network architectures. \salun has large deletion capacity on image and text classification datasets, but much smaller capacity on multimodal tasks, shown in Figures~\ref{fig:cifar100_perf}--\ref{fig:nlvr2_perf}.
%

\begin{figure}[t]
    \centering
    \begin{minipage}{0.45\textwidth}
        \centering
         \includegraphics[width=\textwidth]{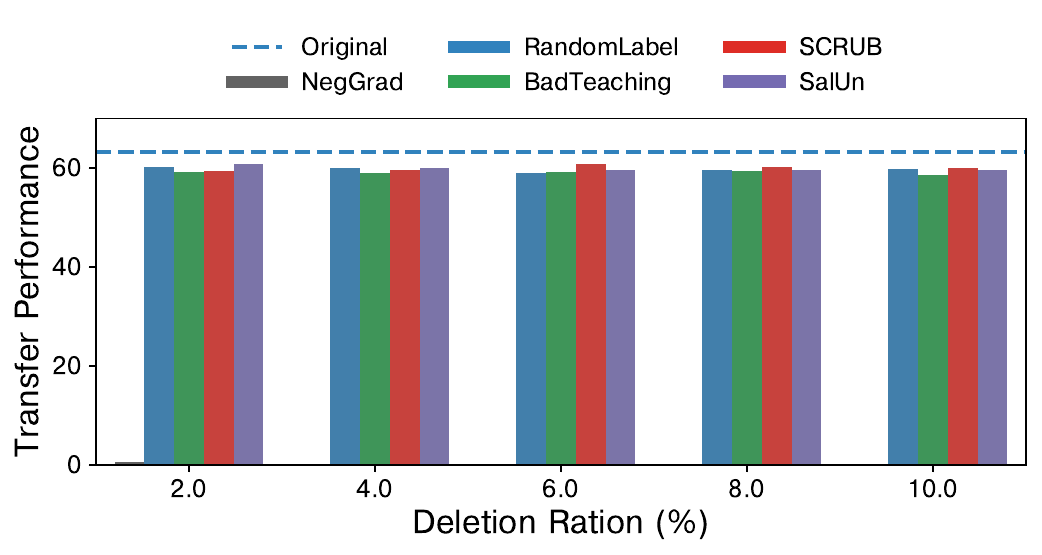}
         \caption{Transfer performances.}
         \label{fig:ood}
    \end{minipage}\hfill
    \begin{minipage}{0.45\textwidth}
        \centering
          \includegraphics[width=\textwidth]{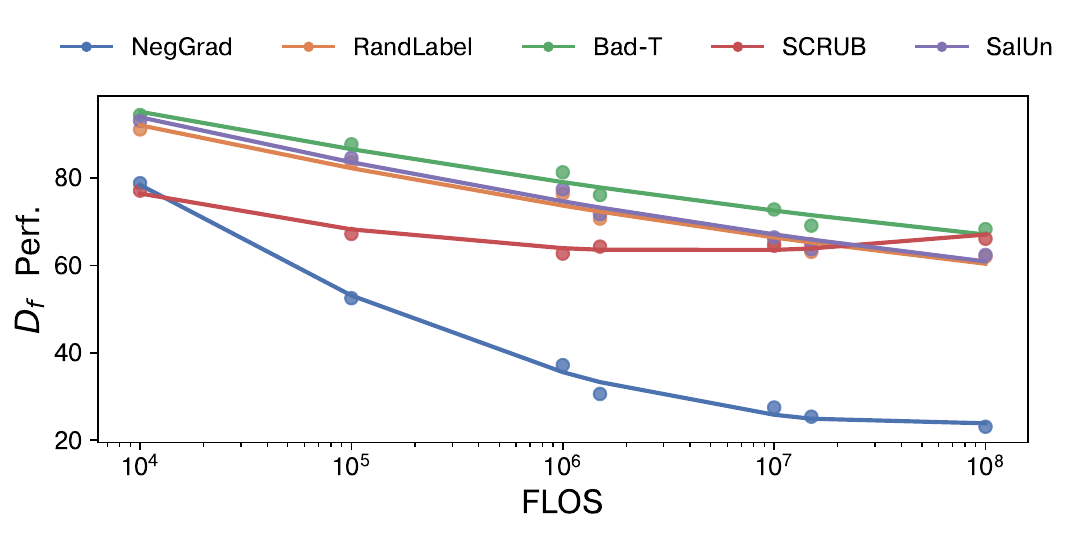}
          \caption{Scaling of \df performance.}
          \label{fig:flops}
    \end{minipage}
    \vspace{-10pt}
\end{figure}

\paragraph{Does unlearning amplify biases?}
A less explored aspect of unlearning in existing works is does MU amplify or restrict the model's dependence on biases in MU. 
To answer these questions, we evaluate the zero-shot transfer performance of $f$ and $f'$ on test examples that are adversarial or from shifted distributions, specifically, CIFAR100-C~\citep{hendrycks2018benchmarking} for CIFAR100, Rotten Tomatoes~\citep{pang-lee-2005-seeing} for IMDB, extra test set from~\citep{suhr-etal-2019-corpus} for NLVR2, UCF101-DS~\citep{robustness2022large} for UCF101, and XSum~\citep{narayan-etal-2018-dont} for SAMSum.
The results show that \neggrad significantly affects models' capability on transfer test sets, while other methods we evaluated do not strongly influence models' dependence on biases, see Figure~\ref{fig:ood}.

\paragraph{Does unlearning follow scaling laws?}
Scaling is a critical aspect to understand the limitations of an unlearning method. 
The results show that \randlabel, \salun, \neggrad, and \badt have a better predictability of performance on \df, given the amount of compute (FLOS), while the performance of \scrub depends on the switch between max steps and min steps. In addition, \neggrad and \scrub have faster speed in decreasing performance on \df. \badt has relatively slower speed, due to the fact that it simultaneously iterate through \df and \dr at every optimization step, which leads to more computing cost than other methods.

\paragraph{Does unlearning benefit from curriculum learning?} The effect of curriculum learning~\citep{bengio_cl,sukhbaatar2018intrinsic} (CL) in MU is an overlooked aspect in existing literature.
MU models often sample batches randomly with no specific order and treat inputs with equal weight. 
We experiment with one common curriculum learning approach SuperLoss~\citep{superloss}, which implements the core principle of curriculum learning. Specifically, it weights training losses based on sample difficulty, weighing down the contribution of samples with large training loss (potentially hard examples) to allow the model to learn from easier samples. As through training, the loss of the hard examples decreases, hard examples are gradually introduced for training. 
The results show that overall SuperLoss results in a slightly larger performance on \df, indicating CL is likely to help model forget less. One exception is that on Speech Commands, CL outperforms Non-CL by 25.4 in accuracy. We defer further experiments with other CL techniques to future work.

\begin{figure}[h]
  \centering
  \includegraphics[width=\textwidth]{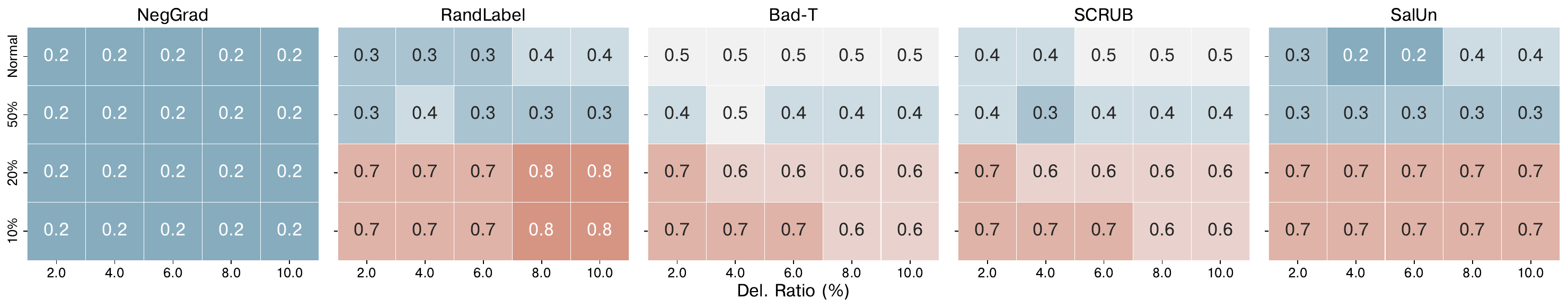}
  \caption{MU training with LoRA.}
  \label{fig:lora}
    \vspace{-10pt}
\end{figure}

\paragraph{Does unlearning benefit from parameter-efficient fine-tuning (PEFT)?} Despite recent advancements of parameter-efficient fine-tuning~\citep{he2022towards}, most MU methods optimize the entire network parameters, which results in significant cost. Only a few approaches have adopted a parameter-efficient strategy~\citep{chen-yang-2023-unlearn,cheng2023gnndelete}. Since PEFT only updates a small portion of the model, it is intuitive to assume that PEFT can maximally retain the knowledge from the original model without compromising unlearning. To validate this hypothesis in the context of MU, we experiment with LoRA~\citep{hu2022lora}.
The results show that most methods can benefit from PEFT, where the performance gap on \df is less than 10 points in accuracy. However, the amount of trainable parameters in MU is much larger than that of fine-tuning. As the trainable parameters are less than 50\% of the original size, the performance on \df is close to that of \dr. Such performance persists even with larger learning rate and longer training time, indicating unlearning \df cannot be achieved below the threshold of 50\%, see Figure~\ref{fig:lora}. This minimum trainable threshold~\citep{hu2022sparse,su-etal-2023-exploring} is much larger than non-MU tasks with as low as a few thousand parameters, since selective knowledge removal is a more challenging task. Meanwhile, the performances on \dt and \dr are not affected, indicating LoRA forgets less and slower in MU.

\paragraph{Which design choices are effective for machine unlearning?}
For discriminitive tasks, corrupting gradient is a less effective approach compared to corrupting data (\randlabel, etc.) and model (\badt). Corrupting gradients can discard learned knowledge and therefore we suggest not using it in isolation without other constraints. However, this approach has a greater potential for generative tasks. 
It is generally more effective to simultaneously iterate through \df and \dr (\badt) or randomly iterate through the training set (\randlabel, \salun), than to clearly separate \df and \dr. For example, \scrub takes a few passes on \df to forget the deletion set before learning on \dr to retain non-deleted data. On the other hand, simultaneous processing of \df and \dr lead to higher computational cost.\looseness-1
Using representation or task loss as a measurement of learned knowledge can adapt to both discriminative and generative tasks, while using logits (\badt) has a much more restricted application merely on classification tasks.\looseness-1

\paragraph{Is a lower performance on $D_f$ always better?} \label{sec:lowdf}
Previous works focus on driving the performance on \df to as low as possible. We suggest that excessively low score on \df might reveal information or indicate its existence, which may be taken advantaged by adversary. Moreover, unlearning does not mean a model should completely lose its capability of handling specific samples in \df. 
Instead, a balanced approach where the unlearned model maintains a reasonably low performance on \df is preferable. Recent works are focusing on this direction, such as zero-retrain evaluation~\citep{Chundawat2022CanBT}, knowledge gap on $D_f | D_T$~\citep{wang-etal-2023-kga,cheng2023gnndelete}. We defer further analysis on the desirable performance of $D_f$ to future work.\looseness-1




\begin{table}[t]
\caption{Contribution of curriculum learning in MU. }
\label{tab:cl}
\centering
\tiny
\begin{tabular}{l|cccccc|ccc|c}
\textbf{Dataset} & CIFAR-100 & IMDB & DDI-2013 & NLVR\textsuperscript{2} & Speech Commands & UCF101 & SAMSum & BioFact & Tiny ImageNet & Ave \\
\toprule
\textbf{Non-CL} & 55.5 & 68.6 & 53.4 & 58.1 & 42.8 & 76.6 & 28.5 & 17.4 & 21.1 & 46.9 \\
\textbf{CL}     & 56.4 & 68.6 & 61.9 & 58.2 & 17.4 & 77.0 & 19.2 & 17.3 & 20.9 & 44.1 \\
\textbf{Gap}    & -0.9 &    0 & -8.5 & -0.1 & 25.4 & -0.4 &  9.3 &  0.1 &  0.2 &  2.8 \\
\bottomrule
\end{tabular}
\end{table}

\section{Conclusion}

\paragraph{Conclusion} We propose \method, the first comprehensive machine unlearning (MU) benchmark that spans various tasks and data modalities, including those previously unexplored in MU. We introduce a unified taxonomy of existing MU works, which highlights their unique design choices and establishes connections between them. We also conduct extensive experiments with commonly-used and recent MU algorithms using \method, discovering that audio and video tasks require more focused development of MU techniques. In addition, we explore several overlooked yet crucial aspects of unlearning, such as bias, parameter-efficiency, curriculum learning, and deletion capacity. Finally, we develop an open-source package of \method to provide unified data
splits, and implement a suite of commonly-used MU methods and their design choices to  enable fast experimentation and fair comparisons across MU methods. The package along with a leaderboard are structured to easily incorporate new datasets and tasks in future.
We will continue to expand \method by incorporating more datasets and tasks.\looseness-1


\paragraph{Future Works}
There are several venues for future work including:
\emph{(a): MU methods for under-investigated tasks and modalities}: existing unlearning methods are primarily developed for text or image data types. Our experiments on \method show that current models severely underperform in audio and video contexts. A promising area of research is to extend MU to these data modalities and tasks through focused development of MU techniques to ensures comprehensive MU capability. 
\emph{(b) Efficient MU methods}: existing unlearning methods require extensive training, either tuning the entire model or training on large portions of the dataset. Meanwhile, most methods do not benefit from PEFT. Future research can focus on developing more efficient MU methods using approaches like zero-shot methods, sparse methods, and curriculum learning methods to speed up the unlearning process.
%
%
\emph{(c) Explainability}: understanding why certain samples are more easily forgotten than others could shed light on inner working of MU methods and improve MU performance.  Therefore, investigating the complexities of samples that affect their retention or deletion is a promising area of research. 
%
\emph{(d) Evaluation}: current evaluation of MU is still in its early stage and demands more attention. Refining current evaluation strategies and metrics will be crucial for advancing the field.
\emph{(e) Theoretical guarantee of MU}: most current non-DP-based MU approaches do not provide theoretical guarantees. A critical future directions is to develop theoretical frameworks that provide bounds performance bounds for MU.\looseness-1

\paragraph{Limitations}
While our work marks significant progress, it has the following limitations:
\emph{(a): Not all MU algorithms are evaluated}: due to the significant cost and resource constraints, we focused on a selection of recent, well-performing and representative approaches rather than an exhaustive examination of all MU models.
\emph{(b): Breadths of experiments}. Our investigation into parameter-efficient fine-tuning and curriculum learning were limited to specific methods like LoRA and SuperLoss, though other more effective approaches exist.
\emph{(c): Not all tasks are included}. There are some relevant tasks that are not currently included in \method, such as those related to graphs, recommendation, or retrieval tasks. We plan to expand the range of tasks and datasets in ongoing development of \method.\looseness-1


\section*{Broader Impact Statement}
Our work lays a foundation for fair and consistent evaluation of machine unlearning techniques and its applications, including the Right To Be Forgotten (RTBF) in AI models, which ensures the protection of personal data and the integrity of AI systems.

\bibliography{reference,anthology1,anthology2}
\bibliographystyle{iclr2025_conference}


\appendix
\section{Appendix}

\subsection{Related work}
\paragraph{Categorization of unlearning methods} \emph{Exact unlearning methods} divide the remaining data into several shards and train a separate model on each subset of data. Then all models are combined to make a prediction. They work under different scenarios, including on images~\citep{bourtoule2021machine,Wu2020DeltaGradRR,pmlr-v119-wu20b,9796721,Dukler_2023_ICCV,Lin_2023_CVPR}, on graphs~\citep{Chen2021GraphU}.
\emph{Differential Privacy-based methods} adopt a one-shot weight update followed by added noise to model weights, whose probability distribution is indistinguishable from that of a model retrained from scratch with theoretical guarantee~\cite{Golatkar2020EternalSO,guo-2020-certified-removal,neel2021descent,pmlr-v139-brophy21a,ceu,izzo2021approximate,suriyakumar2022algorithms,liu2023certified}.
\emph{Teacher-student unlearning methods} formulates unlearning as selectively transferring the knowledge into the unlearned model (student). Usually, the teacher on the non-deleted data is the original model, while the teacher on deleted data is opposite to the original model~\cite{wang-etal-2023-kga,unbound,Chundawat2022CanBT,cheng2023gnndelete,fan2024salun,fast-yet}.

\paragraph{Unlearning for discriminative tasks}
Unlearning works in discriminative tasks covers image classification~\citep{Foster2023,Lin_2023_CVPR,jia2023model,zhang2022prompt}, text classification~\cite{li-liu-2023-make,Mehta_2022_CVPR,Cha_Cho_Hwang_Lee_Moon_Lee_2024,Kang2024}, node / edge classification on graph-structured data~\citep{chen2021graph,Chien2023EfficientMU,pmlr-v206-cong23a,cheng2023gnndelete,ceu,cheng2023gnndelete,sinha2023distill}, regression~\citep{pmlr-v202-tarun23a}, image retrieval~\citep{10.1145/3503161.3548378}, multimodal classification tasks~\citep{cheng2023multimodal,poppi2024safeclip,li2024single}, Bayesian models~\cite{nguyen2020variational}, recommender systems~\cite{10.1145/3485447.3511997,li2022making,li2023ultrare}, k-means~\citep{pan2023machine}, and intelligent agents~\citep{pmlr-v199-liu22a}. 
Many other works focus on class unlearning, i.e. removing all samples with a specific class~\citep{Chen2023BoundaryUR}. However, discriminative tasks on audio and video have been limitedly studied, which this work bridge the gap.

\paragraph{Unlearning for generative tasks}
Unlearning for generation models centers on removing copyrighted, private, NSFW, or biased content from generative models, including diffusion models~\citep{Gandikota_2023_ICCV,Kumari_2023_ICCV,Gandikota_2024_WACV,liu2024implicit,fuchi2024erasing,fan2024salun}, image-to-image models~\citep{li2024machine}, text summarization models~\citep{chen-yang-2023-unlearn}, translation models~\citep{wang-etal-2023-kga}, and text generation models~\citep{lu2022quark,jang-etal-2023-knowledge,kassem-etal-2023-preserving,chen-yang-2023-unlearn}.

\paragraph{Unlearning in LLMs}
Recently, more attention has been paid to unlearning in LLMs. Most works focus on gradient ascent to forget copyrighted content~\citep{eldan2023whos}. \citet{yao2024large} designed two additional losses: 1) predicting if answer is gramatically correct, and 2) maintaining performance. SOUL~\citep{jia2024soul} leverages second-order optimization techniques. Other approaches include sparsity~\cite{learn-to-forget} and operations on gradient~\cite{pmlr-v134-ullah21a,Hoang_2024_WACV}. Applications of unlearning include removing bias~\cite{setlur2022adversarial,chen2023fast}, alleviating backdoor attack~\cite{wei2023shared}, conducting data poinson attack~\cite{di2023hidden}.

\paragraph{Unlearning evaluation}
Evaluation of MU include the effectiveness of exact / DP-based unlearning~\citep{thudi2021necessity}, adversarially trained models~\cite{Liu_2023_ICCV}, adversarially evaluation~\citep{goel2022evaluating}, red-teaming tool for concept removal~\citet{tsai2024ringabell}, verification~\cite{sommer2020towards}, sequential deletion~\citep{NEURIPS2021_87f7ee4f}, vulnerability to attack~\cite{zhao2023static}, trade-off with reverting decisions~\cite{pawelczyk2023on}, different choices of deleted points~\citep{fan2024challenging}, theoretical capacity of deletion~\citep{liu2023certified}, under shallow models~\cite{10.1145/3448016.3457239,ginart2019making}, under zero-shot setting~\cite{chundawat2022zero}. 

\paragraph{Task-specific MU benchmarks}
In general, datasets and benchmarks for unlearning is under-explored. Most works draw samples as deleted data from existing datasets and choose different subsets from paper to paper. UnlearnCanvas is a benchmark for unlearning for diffusion models~\citep{zhang2024unlearncanvas}. TOFU~\citep{maini2024tofu} is a benchmark for unlearning fictitious author profiles in LLMs. Conversely, we test LLMs with unlearning real profiles, as such information appears in the pretraining corpus of the LLMs, which aligns with the unlearning setting.


\subsection{Implementation details}
For all methods, we adopt a batch size of 32 and Adam optimizer. We search for the best learning rate in $[1e-5, 5e-5, 1e-4, 5e-4]$. All experiments are conducted on NVIDIA A100 GPUs.

\subsection{Original models}\label{sec:ori}
We release the following 20 network architectures and 34 different scales to serve as original models in our benchmark.

For CIFAR-100, we train ResNet~\citep{He_2016_CVPR} (18, 34, 50 layers), MobileNet V2~\citep{Sandler_2018_CVPR}, ConvNext~\citep{Liu_2022_CVPR}, ViT~\citep{dosovitskiy2021an} (Base, Large), and Swin-Transformer (Tiny, Base).
For IMDB, we train BERT~\citep{devlin-etal-2019-bert} (base and large), DistilBERT~\citep{sanh2020distilbert}, and Electra~\citep{clark2020electra} (Base).
For DDI, we train BioBERT~\citep{Lee_2019}, PubMedBERT~\citep{Gu_2021} (abstract only and full text).
For NLVR2, we directly take the Vilt~\citep{kim2021vilt} model finetuned on NLVR2 from the original paper.
For Speech Commands, we train HuBERT~\citep{hsu2021hubert} (Base, Large, X-Large), Wav2Vec2.0~\citep{baevski2020wav2vec} (Base, Large), Whisper~\citep{radford2022robust} (Tiny, Base).
For UCF101, we train VideoMAE~\citep{tong2022videomae} (Base, Large).
For SAMSum, we train T5-V1.1~\citep{lester-etal-2021-power} (Small, Base, Large, X-Large).
For Biography, we directly take the instruction tuned Alpaca~\citep{alpaca} (7B, 13B), Vicuna V1.3~\citep{zheng2023judging} (7B, 13B).
For Tiny ImageNet, we directly take the Stable Diffusion V1.4~\citep{rombach2022highresolution} from the original paper.

\subsection{Dataset level performance}
\begin{figure}[h]
  \centering
  \includegraphics[width=\textwidth]{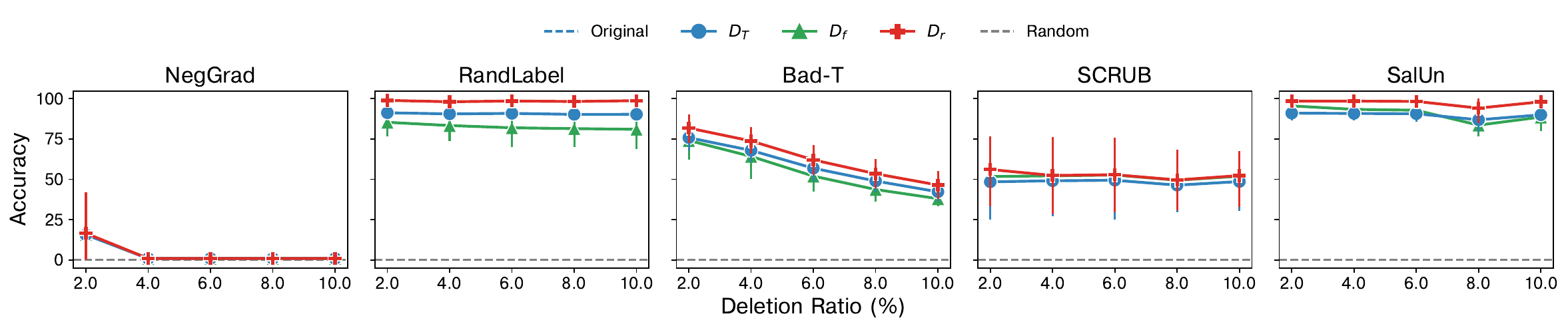}
  \caption{Performance on CIFAR-100.}
  \label{fig:cifar100_perf}
\end{figure}

\begin{figure}[h]
  \centering
  \includegraphics[width=\textwidth]{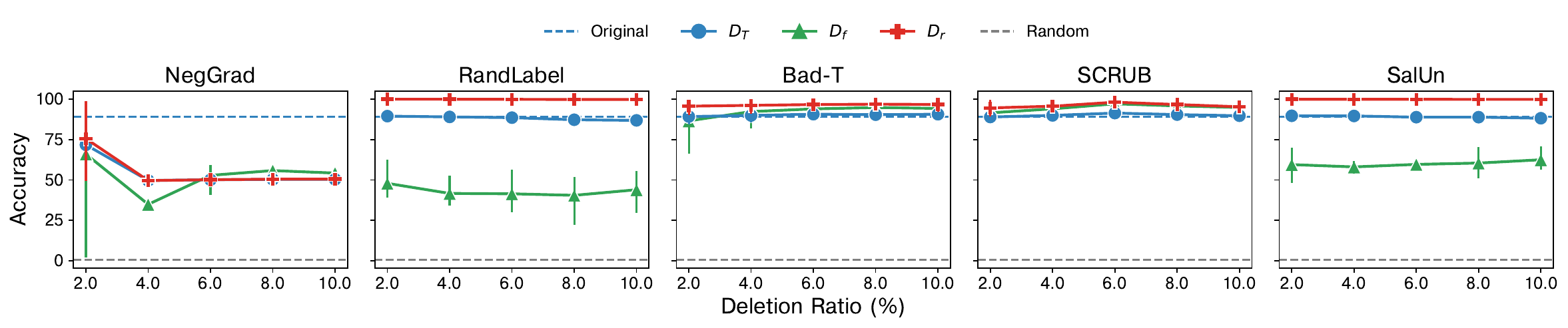}
  \caption{Performance on IMDB.}
  \label{fig:imdb_perf}
\end{figure}

\begin{figure}[h]
  \centering
  \includegraphics[width=\textwidth]{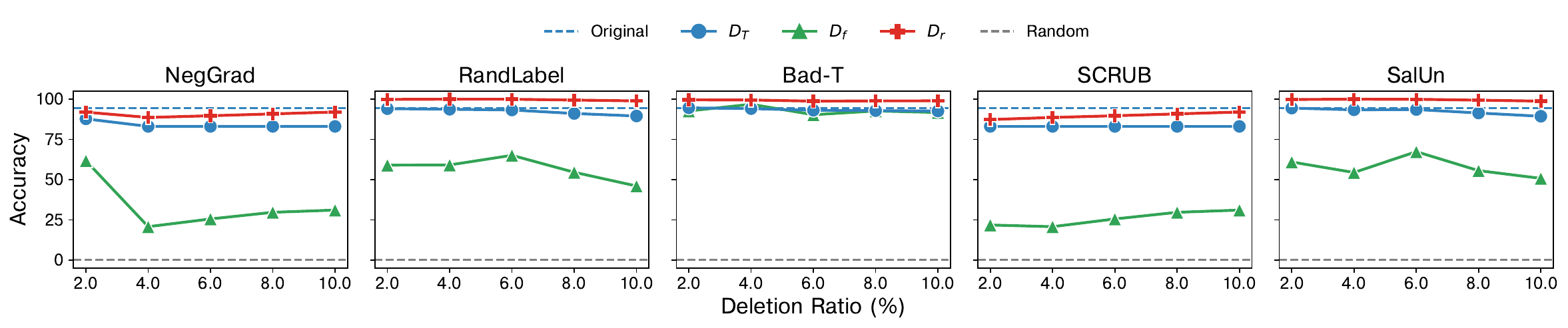}
  \caption{Performance on DDI-2013.}
  \label{fig:ddi_perf}
\end{figure}

\begin{figure}[h]
  \centering
  \includegraphics[width=\textwidth]{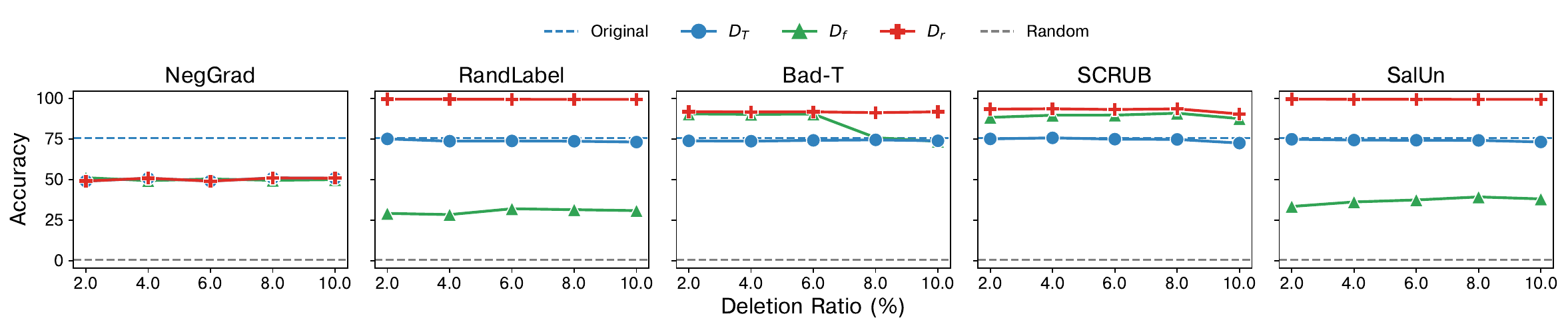}
  \caption{Performance on NLVR\textsuperscript{2}.}
  \label{fig:nlvr2_perf}
\end{figure}

\begin{figure}[h]
  \centering
  \includegraphics[width=\textwidth]{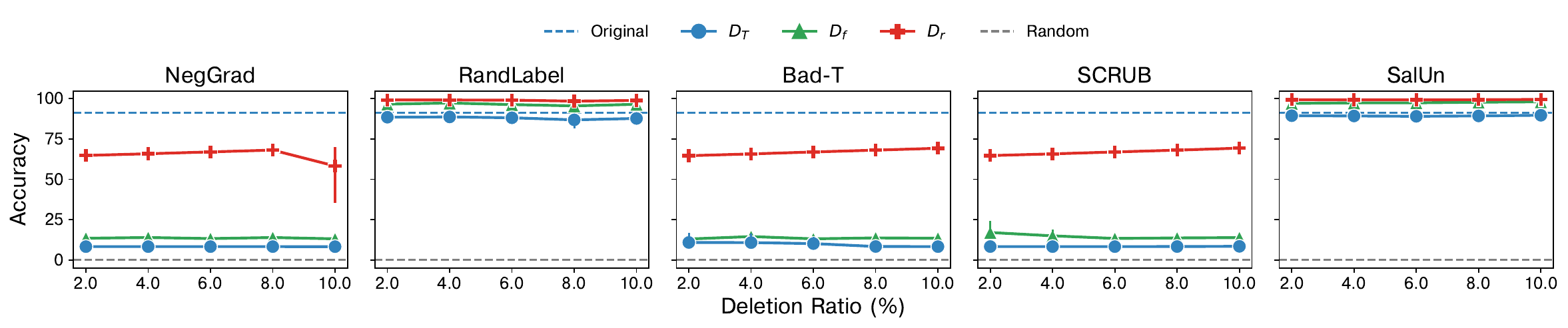}
  \caption{Performance on Speech Commands.}
  \label{fig:superb_ks_perf}
\end{figure}

\begin{figure}[h]
  \centering
  \includegraphics[width=\textwidth]{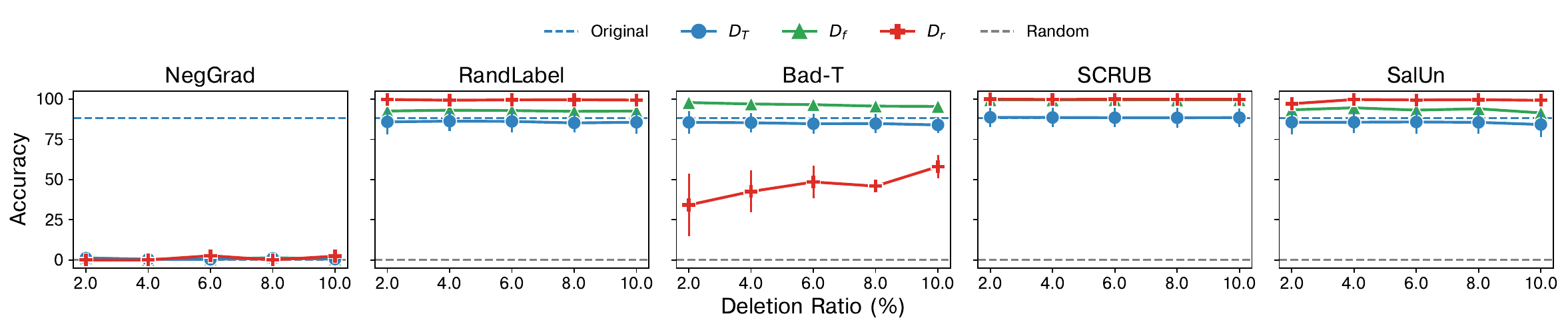}
  \caption{Performance on UCF101.}
  \label{fig:ucf101_perf}
\end{figure}

\begin{figure}[h]
  \centering
  \includegraphics[width=\textwidth]{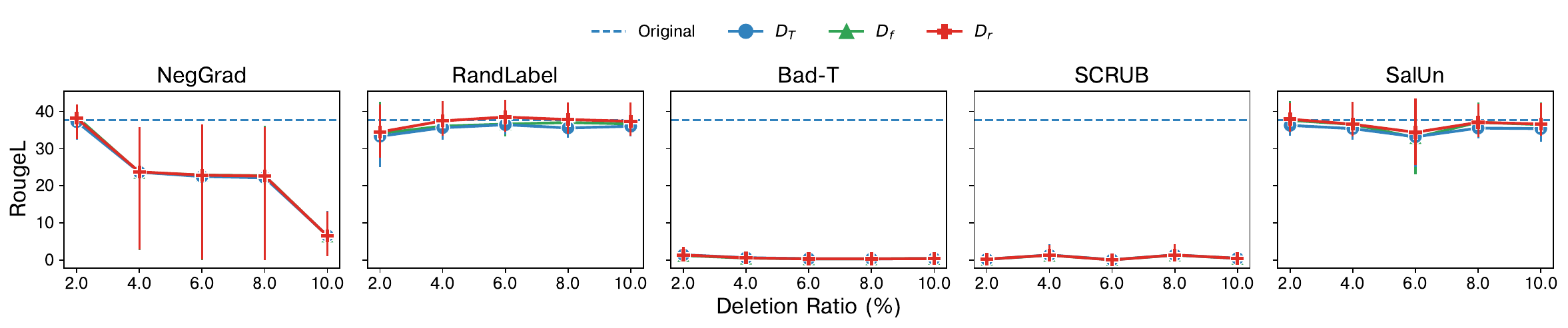}
  \caption{Performance on SAMSum.}
  \label{fig:samsum_perf}
\end{figure}

\begin{figure}[h]
  \centering
  \includegraphics[width=\textwidth]{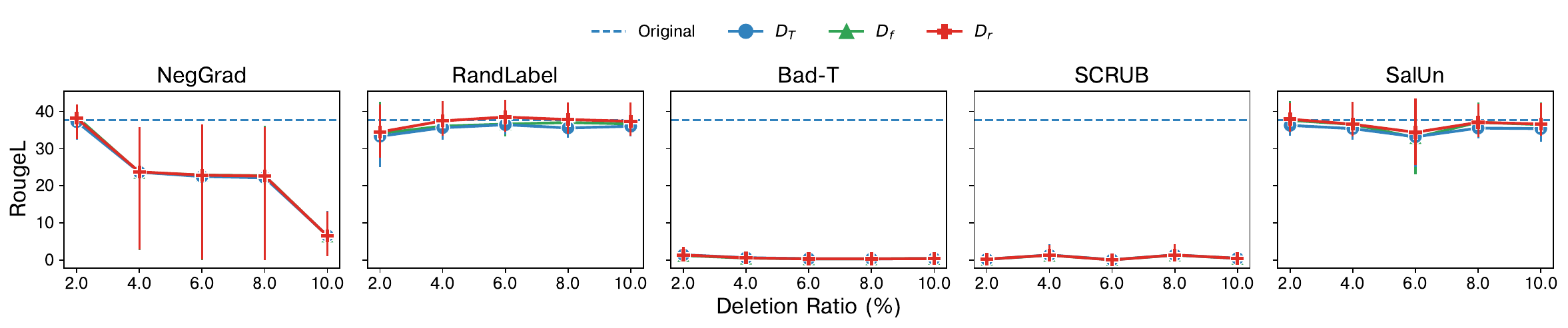}
  \caption{Performance on BioFact.}
  \label{fig:biofact_perf}
\end{figure}

\begin{figure}[h]
  \centering
  \includegraphics[width=\textwidth]{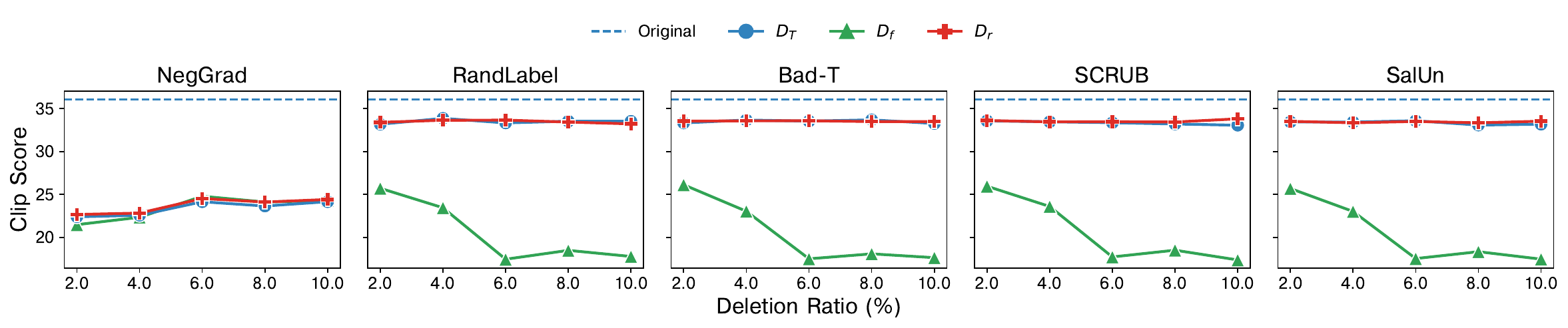}
  \caption{Performance on Tiny Imagenet.}
  \label{fig:tiny_imagenet_perf}
\end{figure}

We present the performance for each dataset in Figure~\ref{fig:cifar100_perf}-\ref{fig:tiny_imagenet_perf}.

\begin{table}[h]
\caption{Accuracy on \df for image generation task.}
\label{tab:image_gen_acc}
\centering
\tiny
\begin{tabular}{l|c}
\textbf{Method} & Acc ($\downarrow$) \\
\toprule
\neggrad    &  3.7 \\
\randlabel  & 64.6 \\
\badt       & 69.1 \\
\scrub      & 75.8 \\
\salun      & 48.2 \\
\end{tabular}
\end{table}

\subsection{More results}
\label{sec:time}
We present the performance on LoRA in Figure~\ref{fig:lora}, membership inference attack in Table~\ref{tab:mi} and unleanring time in Table~\ref{tab:time}.

\begin{table}[h]
\caption{Success rate of membership inference attack.}
\label{tab:mi}
\centering
\tiny
\begin{tabular}{l|c}
\textbf{Method} & Success Rate (\%) ($\downarrow$) \\
\toprule
\neggrad    &  8.6 \\
\randlabel  & 10.7 \\
\badt       & 14.7 \\
\scrub      & 10.8 \\
\salun      & 11.5 \\
\end{tabular}
\end{table}

\begin{table}[h]
\caption{Average unlearning time across all datasets.}
\label{tab:time}
\centering
\tiny
\begin{tabular}{l|c}
\textbf{Method} & Unlearning time (hrs) ($\downarrow$) \\
\toprule
\neggrad    &  8.6 \\
\randlabel  & 10.7 \\
\badt       & 14.7 \\
\scrub      & 10.8 \\
\salun      & 11.5 \\
\end{tabular}
\end{table}

\end{document}